# Comparing learning algorithms in neural network for diagnosing cardiovascular disease

Mirmorsal Madani
Computer Department
Islamic Azad University of Gorgan Branch, kordkuy Center
mt_madani@yahoo.com

*Abstract*— Today data mining techniques are exploited in medical science for diagnosing, overcoming and treating diseases. Neural network is one of the techniques which are widely used for diagnosis in medical field. In this article efficiency of nine algorithms, which are basis of neural network learning in diagnosing cardiovascular diseases, will be assessed. Algorithms are assessed in terms of accuracy, sensitivity, transparency, AROC and convergence rate by means of 10 fold cross validation. The results suggest that in training phase, Lonberg-M algorithm has the best efficiency in terms of all metrics, algorithm OSS has maximum accuracy in testing phase, algorithm SCG has the maximum transparency and algorithm CGB has the maximum sensitivity.

## I. INTRODUCTION

Cardiovascular disease is any kind of disease which influences on circulatory system. It mainly includes heart diseases, cerebrovascular diseases, kidney and arterial diseases. According to statistics in 2006, 26% of death rates in the USA are because of heart diseases [3]. Data mining in medical science has been used very much and in recent years it has been widely studied. A wide range of problems in medical science is associated with diagnosis and they are solved through various experiments. From view of data mining, prediction in diagnosis is among data classification problems. Classification includes studying features of a new object and allocating it to one of pre-determined sets.

Neural network has obtained significant importance as a classification technique in recent years among pattern classification algorithm and machinery learning. Neural network is preferred to other methods in terms of its high acceleration, accuracy and efficiency during colliding with large data basis. Learning is a key ability of neural network. Learning rules are algorithms for finding suitable weights or other parameters of network. There are various algorithms for training neural network. It is a hard task to select an appropriate learning algorithm for neural network and it depends on many factors.

In this article we seek to review efficiency of 9 algorithms of neural network learning for diagnosis and classification of those who suffer from heart diseases. Efficiency of the algorithms will be reviewed in testing and training phases in terms of accuracy, sensitivity, transparency, and AROC and convergence rate. In the first section we describe neural network. In the second section various algorithms of neural network learning are briefly explained. In the third section experiment results are explained and in the end of the article conclusion and future works are presented.

## II. NEURAL NETWORK

Artificial neural network originated from biological systems [5]. Neural network was composed of too many neurons and it has the ability of learning from samples, such as human brain. Neural network can do tasks that cannot be performed by means of linear planning. In the network information is available in communication between neurons directly. The data were obtained from biological systems through learning [5]. The data which are achieved through learning can be applied for decision-making on new samples. Multi-layered perceptron neural network includes one input layers, hidden layer and output layer. In some articles efficiency of neural network was widely approved in diagnosing various diseases, such as skin disease [1], oncology [4], and radiology [2] and so on. According to figure 1, neural network was used with three layers, 13 input layers in the first layer, 7 neurons in the hidden layer and one neuron in output layer. Sigmoid transfer function was used in the hidden and output layers.

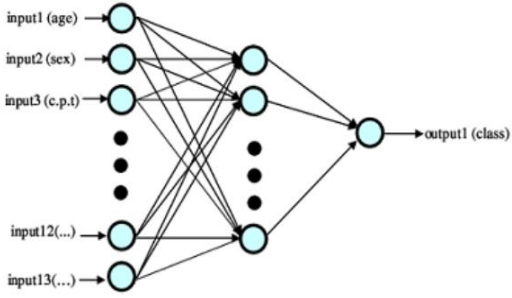

Fig 1. neural network presented for diagnosis

TABLE1. gradient reduction algorithms

| Algorithm | Acronyms |
|---|---|
| Lonberg-M | LM |
| Newton-Gaussian | BFG |
| Reactionary propagation | RP |
| Scale combined gradient | SCG |
| Gradient reduction with Powell/Beale Restarts | CGB |
| Reduction gradient with Fletcher-powell | CGF |
| Polak-Ribiere gradient reduction | CGP |
| One-step-secant | OSS |
| Variable learning rate | GDX |

### III. NEURAL NETWORK TRAINING ALGORITHMS

Neural network design is highly dependent on type of neural network learning algorithm. Leaning algorithms are used for obtaining optimum parameters by means of efficiency function (mean square error). The algorithms are mainly located in two groups: first-time algorithms (based on gradient reduction methods) and second-time algorithm. In table 1, nine algorithms with neural network training basis along with acronym of each algorithm were presented. 9 algorithms will then be described briefly [5].

- *Gradient reduction algorithms*

This algorithm is a first-time algorithm and uses the first derivate of the error function for finding minimum in error space. According to equation (1), gradient g can be defined in form of first-time derivative from E total error function.

$$g = \frac{\partial E(f,x)}{\partial W} = \left[\frac{\partial E}{\partial w_1} \quad \frac{\partial E}{\partial xw_1} \cdots \frac{\partial E}{\partial w_N}\right]^T \quad (1)$$

Defining gradient g, rules of updating gradient reduction algorithm can be written as equation (2).

$$w_{k+1} = w_k - \alpha_k\, g_k \quad (2)$$

- *GDX variable learning rate*

In standard gradient reduction methods, LR rate is constant in all the training phases, while algorithm efficiency is highly dependent on LR rate. With large LR value, the algorithm has many fluctuations and with small LR value the algorithm needs much time for convergence. It is recommended to change LR level during training process based on level of efficiency in order to improve efficiency of algorithm. Size of a flexible LR should make sufficient stability for algorithm [5].

- *Reactionary propagation (RP)*

Multi-layered networks often use sigmoid transfer function in their hidden layer. The functions compress inputs, which are in maximum range, into a small range. Generally, in these functions gradient becomes zero if the input is large. This leads to a problem in gradient reduction algorithm, since at this point the gradient will become less and makes some quantitative changes in weights and biases. As a result weights and biases will remain far from their optimum value. The aim of reactionary propagation algorithms is to remove side effects in minor derivatives. Only derived mark is used for specifying update direction of weights. Size of derivative will leave no effect on weight`s updating. Changes of weight can be determined by means of a single updating value. Weight update and bias values will increase if derivative value does not change its signal in sequenced repetition. Also updating value will be reduced with one factor where derivative changes its mark to previous repetition. Updating value does not change with zero derivatives. With fluctuation of weights, weights will change in small amount. If weight keeps changing with the same direction for several repetitions, value of change will increase. The algorithm has high efficiency to standard algorithm of gradient reduction. Also, the algorithm needs less memory [4].

- *Combined gradient algorithms*

In most algorithms, learning rate is used for specifying size of steps in updating weights. Size of each step is adjusted for each repetition in most of the algorithms. Hence, a searching operation is conducted between all the gradients where efficiency function is minimized along the line. In equation (3) all the algorithms start searching in the first repetition for maximum gradient reduction.

In equation (4) a linear search was conducted to determine optimum distance with linear search.

$$p_0 = -g_0 \quad (3)$$

$$w_{k+1} = w_k + a_k p_k \quad (4)$$

Then, direction of next search is accompanied with previous direction. According to equation (5), general procedure for specifying a new search direction is a combination of new search direction and the previous one.

Different accompanied gradients are separated for calculating $P_k$ by means of their behavior. For Fletcher-Reeves, $P_k$ is calculated by means of equation (6).

$$p_k = -g_k + \beta_k p_{k-1} \quad (5)$$

$$\beta_k = \frac{g_k^T g_k}{g_{k-1}^T g_{k-1}} \quad (6)$$

Combined gradient algorithms are very quick and sometimes they are even quicker than reactionary propagation. Of course every problem has different results. These algorithms need more memory compared to simple algorithms.

Polak-Ribiere is another algorithm in which parameters are calculated in terms of equations (7) and

$$p_k = -g_k + \beta_k p_{k-1}$$
$$\beta_k = \frac{\Delta g_k^T g_k}{g_{k-1}^T g_{k-1}} \quad (7,8)$$

In all the algorithms, search direction is reset in certain repetitions to negative gradient. Standard reset point is where number of repetitions is equal with number of network parameters. But there are some methods for specifying these points which increase efficiency. In Powell-Beale Restars algorithm, when there is not enough balance between current and previous gradients, reset operation is conducted for search. The problem is checked out in equation (9).

$$w_{k+1} = w_k - H_k^{-1} g_k \quad (9)$$

If above condition exists, search direction is reset. The algorithm has more efficiency in some problems compared to CGP. But it is hard to comment for every problem. In return the memory level used in this method is more than CGF [5].

- *Scale combined gradient*

All the algorithms discussed so fare regarding combined gradient require a linear search. The linear search is expensive to be calculated, since network should react to all the training inputs and calculate several parameters for each search. Scale combined gradient algorithm is designed in a way that it does not need any time-consuming linear search. The algorithm is very complex and cannot be explained here. But it is based on combining two methods of combined gradient and Lonberg-M. The algorithm needs more repetition for convergence compared to other combined gradient algorithms. But calculation in each repetition is reduced significantly. Since linear search was not conducted in this method. Memory space needed in this algorithm is similar to Fletcher-Reeves [5].

- *Quasi-Newton algorithm*

In this method all the elements of $g_1, g_2 \ldots g_N$ are functions of weights and all the weights are independent of each other in terms of being linear. Therefore, updating rules for Newton method is calculated by means of equation 10.

$$H = \begin{bmatrix} \frac{\partial^2 E}{\partial w_1^2} & \frac{\partial^2 E}{\partial w_1 \partial w_2} & \cdots & \frac{\partial^2 E}{\partial w_1 \partial w_N} \\ \frac{\partial^2 E}{\partial w_2 \partial w_1} & \frac{\partial^2 E}{\partial w_2^2} & \cdots & \frac{\partial^2 E}{\partial w_2 \partial w_N} \\ \vdots & & & \\ \frac{\partial^2 E}{\partial w_N \partial w_1} & \frac{\partial^2 E}{\partial w_N \partial w_2} & \cdots & \frac{\partial^2 E}{\partial w_N^2} \end{bmatrix} \quad (10)$$

In equation (10), H is Matrix Hessian.

$$|g_{k-1}^T g_k| \geq 0.2 \|g_k\|^2 \quad (11)$$

Newton methods usually have more suitable and quicker convergence to gradient reduction algorithms. But this improvement happens only when second class approximation of error function is logical. Otherwise, the algorithm becomes divergent. In order to obtain Matrix Hessian H, second derivative of total error function should be calculated and this could be very complex and expensive to calculate. As a result they are not suitable for neural network. In quasi-Newton method which is a Newton-based method algorithm, the second derivative is not required to be calculated and it needs less calculation costs. In this method Jakobian matrix was applied instead of Matrix Hessian. In Newton-Gaussian algorithm, updating rules are calculated by means of equation (12).

$$H \approx J^T J + \mu I \quad (12)$$

In this equation e and J is Jakobian matrix. Newton-Gaussian algorithm is still facing convergence difficulties such as Newton algorithm for optimizing complex error level [4].

- *Lonberg-M algorithms*

This algorithm belongs to gradient reduction algorithm and Newton-Gaussian. Fortunately, this algorithm inherits convergence rate of Newton-Gaussian algorithm and consistency of gradient reduction methods. Although Lonberg-M algorithm has slower convergence than Newton-Gaussian algorithms, it has quicker convergence than gradient reduction methods. In order to make sure that Hessian matrix $J^TJ$ is inversion allowed, algorithm Lonberg-M introduces estimation for Hessian matrix in terms of equation (13):

$$w_{k+1} = w_k - (J_k^T J_k + \mu I)^{-1} J_k e_k \qquad (13)$$

Where μ is always positive which is called combination coefficient. And I am identity matrix. It can be learnt from equation (13) that elements of main diameter in Hessian matrix is more than zero. Therefore, this makes matrix H to be inversion allowed all the time.

Rules of updating Lonberg-M algorithm can be calculated by means of formula (14).

$$w_{k+1} = w_k - (J_k^T J_k)^{-1} J_k e_k \qquad (14)$$

Combining two algorithms of gradient reduction and Newton-Gaussian, algorithm Lonberg-M is awitched between the two algorithms during leaning process. Once combination coefficient U becomes very small, Newton-Gaussian algorithm is used. When the coefficient is very big, gradient reduction algorithm is used [6].

- *Algorithm BFGS*

This algorithm needs more calculations and space compared to other combined gradient methods. Approximation matrix is hessian n*n where *n* is equal to weights and network bias. High size of matrix makes some problems for storing in terms of level of memory. Therefore, very big networks are recommended to use Rprop methods or combining gradient rather than this method. For smaller networks the algorithm can have better efficiency to other methods [3].

- *Algorithm One Step Secant*

Since BFGS needs too much space and many calculations, another quasi-Newton method was founded with less space and calculations. Algorithm OSS is in fact a bridge between combining gradient and quasi-Newton. The algorithm does not save full hessian matrix, it supposes that previous hessian matrix is valid in each repletion and this led to reduction of calculation and space used [3].

## IV. EXPERIMENTS

It is a difficult task to specify which of the algorithms are suitable for neural network training for classifying medical data, since this problem depends on many factors such as problem complexity, number of training data, data quality, weights and bias in network and so on. To review efficiency of neural network learning algorithms, we may use previous data available in University of California at Irvine (UCI) [4]. We need to normalize inputs and objective before neural network training so that they are scaled in a certain range. In these experiments, data applied are normalized in ranges _1 and 1. The algorithms are assessed in terms of accuracy, sensitivity, transparency, AROC and convergence rate by means of Cross validation 10 fold.

TABLE 2. AVERAGE ACCURACY, SENSITIVITY AND TRANSPARENCY OF GRADIENT REDUCTION ALGORITHMS

| Algorithm | Train set | | | Test set | | |
|---|---|---|---|---|---|---|
| | SEN | SPE | ACC | SEN | SPE | ACC |
| SCG | 0.8671 | 0.7513 | **0.8156** | 0.7665 | 0.7217 | 0.7500 |
| RP | 0.8393 | 0.7238 | **0.7881** | 0.8330 | 0.6234 | 0.7407 |
| OSS | 0.8747 | 0.7658 | **0.8263** | 0.8346 | 0.6997 | 0.7778 |
| LM | 0.9651 | 0.9452 | **0.9564** | 0.8010 | 0.6948 | 0.7519 |
| GDX | 0.7234 | 0.4613 | **0.6070** | 0.7287 | 0.4525 | 0.6148 |
| CGP | 0.8830 | 0.6352 | **0.7720** | 0.8805 | 0.5919 | 0.7519 |
| CGF | 0.8617 | 0.6744 | **0.7786** | 0.8072 | 0.6446 | 0.7296 |
| CGB | 0.8869 | 0.6420 | **0.7770** | 0.8367 | 0.5842 | 0.7333 |
| BFG | 0.8991 | 0.6013 | **0.7671** | 0.8867 | 0.5651 | 0.7333 |

In figure 2, values of AROC exist in testing and training phases for cardiovascular data. In testing algorithm, OSS achieved 0.8378 and in training algorithm LM received 0.9472 which are maximum level of AROC.

Speed of neural network training algorithms are presented in figure 3. Objective function was mean square error and it sought to reduce MSE in repetition of neural training algorithms, in order to reach the value required. According to figure 3, error of algorithm LM is reduced with the most speed than other algorithms.

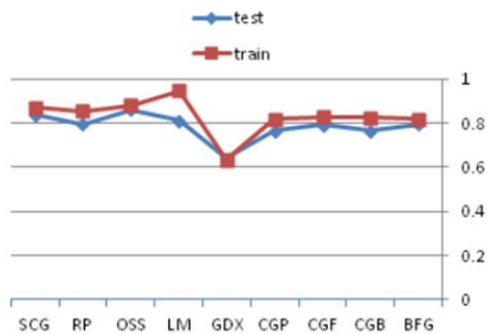

Figure 2. values of AROC in testing phase and training for cardiovascular data

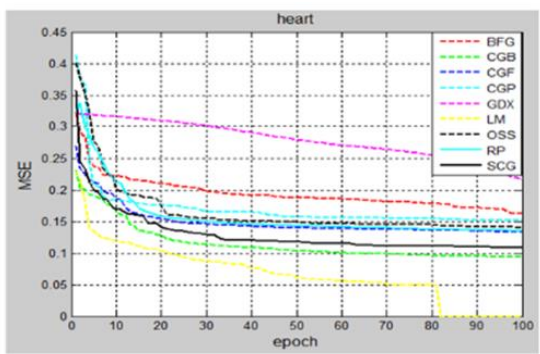

Figure 3. comparing of convergence rate of neural network training algorithms for those suffering from cardiovascular diseases.

## V. CONCLUSION

In this paper, 9 algorithms of neural network learning basis were reviewed for recognizing heart diseases. Neural network with architectural record 1_7_13 was applied (13 neurons in input layer, 7 neurons in hidden layer, and 1 neuron in output layer). In neural network training, algorithm Lonberg-M is in the first place with maximum accuracy, sensitivity, transparency, and AROC and convergence rate. Algorithm OSS has 77.78 accuracy, SCG has 72.17 transparencies and CGB has 83.67 sensitivity which all have maximum average in testing phase. According to the results, neural network has great accuracy in diagnosing those suffering from cardiovascular diseases. This is due to wide communication between non-linear elements which makes interpretation hard for people. As a result, in future works process of discovering rules could be reviewed.